\documentclass[sigconf,authorversion]{acmart}

\AtBeginDocument{%
  \providecommand\BibTeX{{%
    \normalfont B\kern-0.5em{\scshape i\kern-0.25em b}\kern-0.8em\TeX}}}

\copyrightyear{2020} 
\acmYear{2020} 
\setcopyright{acmcopyright}\acmConference[ICMI '20 Companion]{Companion Publication of the 2020 International Conference on Multimodal Interaction}{October 25--29, 2020}{Virtual event, Netherlands}
\acmBooktitle{Companion Publication of the 2020 International Conference on Multimodal Interaction (ICMI '20 Companion), October 25--29, 2020, Virtual event, Netherlands}
\acmPrice{15.00}
\acmDOI{10.1145/3395035.3425228}
\acmISBN{978-1-4503-8002-7/20/10}

\newcommand{\empdial}{}    
\def\empdial/{\textsc{EmpatheticDialogues}}

\newcommand{\pec}{}
\def\pec/{\textit{PEC}}

\newcommand{\emppec}{}
\def\emppec/{\textit{EmpPEC}}

\newcommand{\nonemppec}{}
\def\nonemppec/{\textit{NonEmpPEC}}

\newcommand{\hlrpmi}{}
\def\hlrpmi/{\texttt{HighLowRolePlayMI}}

\newcommand{\hrpmi}{}
\def\hrpmi/{\texttt{HighRolePlayMI}}

\newcommand{\lrpmi}{}
\def\lrpmi/{\texttt{LowRolePlayMI}}

\newcommand{\detector}{}
\def\detector/{$cls^{dct}_{emp}$}

\newcommand{\labeller}{}
\def\labeller/{$cls^{label}_{emp}$}

\begin{document}

\fancyhead{}

\title{Towards Detecting Need for Empathetic Response in\\ Motivational Interviewing}

\author{Zixiu Wu}
\email{zixiu.wu@philips.com}
\affiliation{%
  \institution{Philips Research}
  \city{Eindhoven}
  \country{Netherlands}
}
\affiliation{%
  \institution{University of Cagliari}
  \city{Cagliari}
  \country{Italy}
}

\author{Rim Helaoui}
\email{rim.helaoui@philips.com}
\affiliation{%
  \institution{Philips Research}
  \city{Eindhoven}
  \country{Netherlands}
}

\author{Vivek Kumar}
\authornote{Currently (September 2020) his secondment at Philips Research is in progress.}
\email{vivek.kumar@unica.it}
\author{Diego Reforgiato Recupero}
\email{diego.reforgiato@unica.it}
\author{Daniele Riboni}
\email{riboni@unica.it}
\affiliation{%
  \institution{University of Cagliari}
  \city{Cagliari}
  \country{Italy}}

\renewcommand{\shortauthors}{Wu, et al.}

\begin{abstract}
Empathetic response from the therapist is key to the success of clinical psychotherapy, especially motivational interviewing. Previous work on computational modelling of empathy in motivational interviewing has focused on offline, session-level assessment of therapist empathy, where empathy captures all efforts that the therapist	makes to understand	the	client’s	perspective	and	convey that understanding to the client. In this position paper, we propose a novel task of turn-level detection of client need for empathy. Concretely, we propose to leverage pre-trained language models and empathy-related general conversation corpora in a unique labeller-detector framework, where the labeller automatically annotates a motivational interviewing conversation corpus with empathy labels to train the detector that determines the need for therapist empathy. We also lay out our strategies of extending the detector with additional-input and multi-task setups to improve its detection and explainability.
\end{abstract}



\maketitle

\section{Introduction}
\label{sec:introduction}
Empathy from the counsellor side is widely recognised as essential to building counsellor-client rapport in psychotherapy~\citep{empathy-psychotherapy}\footnote{\url{https://www.stephenrollnick.com/three-pieces-on-empathy/}}. Its importance is particularly acknowledged in motivational interviewing~\citep{mi} (MI), a psychotherapeutic technique that has proved successful in helping people achieve positive behaviour change by eliciting their own motivation. It is also a key aspect of psychotherapeutic interview quality. For example, according to the most used Motivational Interviewing Coding system (MITI~\citep{miti}), assessing MI integrity goes through four dimensions, namely cultivating change talk, softening sustain	talk, partnership, and \textbf{empathy}\footnote{\url{https://casaa.unm.edu/download/miti4\_2.pdf}}.

While such assessments are mostly done by experts manually, recent research has explored automatic analysis and rating of therapist empathy in MI based on text~\citep{n-gram-empathy, psycholinguistic-norms-empathy, deep-learning-empathy-modelling}, speech~\citep{prosody-for-therapist-empathy, speech-rate-entrainment-empathy}, or both~\citep{rate-my-therapist-with-asr}. Nevertheless, these studies are limited since 1) they only evaluate offline, session-level counsellor empathy, instead of offering real-time advice; 2) they nearly all rely on classical machine learning methods with heavy feature engineering, yet pre-trained language models such as BERT~\citep{bert} have not been explored despite their superior results in Natural Language Understanding tasks; 3) they were conducted on undisclosed datasets of MI conversations, which makes their results difficult to verify or replicate.

To address the limitations above, we introduce a novel text-based task of \textbf{detecting} the need for empathy in immediate, turn-level MI counsellor response, and plan to tackle it by fine-tuning large-scale pre-trained language models on publicly available free-access datasets of general and therapeutic dialogues. This new task can be highly valuable for educating inexperienced coaches and providing them with real-time guidance in their first actual sessions.

We will approach the task based on two recent empathy-related dialogue datasets: \textbf{\pec/}~\citep{pec} (Persona-based Empathetic Conversation) and \textbf{\hlrpmi/} ~\citep{high-low-quality-mi}. The former comprises general conversations from Reddit\footnote{Reddit (\url{https://www.reddit.com/}) is an online platform consisting of a number of subforums, a.k.a. \textbf{subreddits}, each corresponding to a specific topic for Reddit users to discuss.} with each annotated as empathetic or non-empathetic at dialogue-level but not turn-level. The latter consists of role-play demonstrations of high- and low-quality MI counselling from YouTube, where each conversation (transcribed) has only session-level ``high'' and ``low'' quality labels. We make the following assumptions about these datasets:
\begin{enumerate}
    \item In the high-quality conversations in \hlrpmi/ (aliased as \hrpmi/), the therapist only shows empathy when the client needs it.
    \item Responses from the listener are predominantly empathetic in the empathetic \pec/ conversations and non-empathetic in the non-empathetic ones.
    \item How the skilled therapist in \hrpmi/ decides whether to show empathy to the client based on the conversation flow is different from the manner in which the ordinary Reddit listener in the empathetic \pec/ conversations makes this decision before reacting to the speaker.
    \item How the therapist shows empathy \hrpmi/ is similar to the way the listener shows empathy in the empathetic \pec/ conversations, meaning that the empathetic responses in two different domains (therapeutic and general conversations) share similar patterns.
\end{enumerate}

Assumption (1) is highly reliable because of the importance of therapist empathy in high-quality MI counselling. Assumption (2) is verified statistically by~\citet{pec} with ``substantial'' inter-annotator agreement. Assumption (3) is also very likely to be true since skilled therapists have been specially trained to engage their clients with empathy, whereas ordinary Reddit users have not. Assumption (4) is reasonable but yet to be substantiated, and we therefore will explore its reliability in our experiments. Enabled by these assumptions, our approach consists of two steps: \textbf{empathy labelling} of \hrpmi/ exploiting \pec/ and \textbf{need-for-empathy detection} using \hrpmi/.

For empathy labelling, we will annotate each therapist utterance in \hrpmi/ with a binary empathy-or-not label. Specifically, we will train a binary classifier (labeller) to distinguish between responses from the empathetic conversations and those from the non-empathetic ones in \pec/, so that the resultant model is able to differentiate between empathetic and non-empathetic responses, based on assumption (2). The labeller will then label each therapist response in \hrpmi/ as empathetic or non-empathetic, based on assumption (4). Note that due to assumptions (1) and especially (3), we cannot directly train a detector on the empathetic dialogues of \pec/ (general conversation domain) to accomplish the need-for-empathy detection task on \hrpmi/ (therapeutic dialogue domain).

If the performance evaluation of the labeller on \hrpmi/ reaches satisfactory levels (e.g. w.r.t. accuracy/F1), we will move on to the proposed need-for-empathy detection task. More concretely, we will train a detector on \hrpmi/ which is now annotated with an empathy label for each therapist utterance. Given the conversation history where the last turn is from the client, the detector will reach a yes/no answer on whether empathy is needed in the immediate therapist response to the latest client utterance.

Both the labeller and the detector will be based on pre-trained language models and trained end-to-end. We will also experiment with additional-input regimes and multi-task designs to enhance the ability of the detector and increase its explainability.

In summary, our contributions are as follows:
\begin{itemize}
    \item We propose a new need-for-empathy detection task, which aims to provide the human therapist with real-time advice on whether to show empathy in their response to the client.
    \item We propose an automatic empathy labelling method that could annotate therapist empathy, using publicly available free-access datasets of empathy-related general dialogues.
    \item We present our plan for approaching the detection task, leveraging the aforementioned automatic annotation and simply requiring free-access MI dialogues in the public domain without manually created empathy labels.
    \item We detail our proposed additional-input and multi-task detector extensions for explainability improvement.
\end{itemize}

\section{Related Work}
\label{sec:related-work}
The work we propose is closely related to prior work on 1) data-driven analysis of therapeutic empathy in MI and 2) text-based approaches to computational empathy in general conversation, as we will address the former using techniques inspired by the latter.

\subsection{Data-Driven Approaches to Therapeutic Empathy Analysis for MI}
Empathy is fundamental in coaching in general and in clinical counselling in particular, such as evidence-based behavioural treatments like MI. Prior work studying the link between empathy and MI delivery has focused on the speech and language of the therapist.

An early text-based attempt~\citep{n-gram-empathy} adopted an n-gram language model for binary classification of utterance empathy.~\citet{psycholinguistic-norms-empathy} used generalised linear regression on other linguistic features as well as psycholinguistic norm features. More recently, long short-term memory networks (LSTMs)~\citep{lstm} were used for turn-level encoding to infer utterance-level behaviours that are further used by a deep neural network (DNN) to predict session-level empathy~\citep{deep-learning-empathy-modelling}.

The other line of work uses speech features.~\citet{prosody-for-therapist-empathy} extracted prosodic features such as pitch and energy from speech signals for empathy classification using linear Support Vector Machines. Speech rate entrainment during dyadic interactions has also been investigated for its relationship to therapist empathy~\citep{speech-rate-entrainment-empathy}. Finally,~\citet{random-forest-counselling-empathy} trained a random forest on both linguistic and acoustic features to identify therapist empathy.

Note that previous work, as listed above, has all targeted offline therapist empathy assessment or identification of contributing factors to high-quality therapy. Our goal, in contrast, is to \textbf{detect} the need for therapist empathy, which makes our proposal closer to the work of~\citet{text-based-behaviour-coding-forecasting}, where a forecaster of MI behavioural codes was developed to provide real-time guidance for the therapist on what action (e.g. reflection/question/$\cdots$) to take next. Our work differs from it in that we take an empathy-centred perspective.

It is worth mentioning that there is more recent work on deep-learning-powered MI analysis, including text-based~\citep{text-based-behaviour-coding-forecasting, comparison-nlp-for-mi-coding, mi-coding-rnn, multi-label-multi-task-behavioural-coding}, speech-based~\citep{speech-based-behaviour-coding}, and multimodal~\citep{multimodal-misc-coding-class-confusions} coding of therapist actions according to MI behaviour codes, but we do not consider this sphere of research since it is still predominantly for offline assessment of therapy quality and does not involve explicit empathy modelling.

\subsection{Text-Based Approaches to Computational Empathy in General Conversation}
Substantial progress has been achieved in text-based sentiment analysis (SA)~\cite{Dridi20192045,Recupero2015211,Recupero2014245,ReforgiatoRecupero20143} in recent years, exemplified by the over-97\% accuracies from the top entries in the SST-2 SA task on the GLUE~\citep{glue} leaderboard\footnote{\url{https://gluebenchmark.com/leaderboard}}. In comparison, text-based empathy analysis, especially for general conversation, attracted less attention, until \empdial/~\citep{facebook-empathy}, a dataset of empathetic conversations grounded in emotions and situations, was made public. Since then, various studies have explored creating an empathetic open-domain conversation agent. Therefore, we focus on the approaches to computational empathy for such agents in this section.

Many of those agents incorporate \textbf{current} user emotion during their response generation. An early study~\citep{mojitalk} used emotion labels (emojis in tweets) as extra input to train a conditional variational encoder for response generation. Similarly,~\citet{emotion-context-vector} created an user-emotion context vector in addition to the HRED~\citep{hred} response-generation framework. In~\citep{facebook-empathy}, the emotion of the user is categorised as an emotion label to prepend each user utterance so that the response generator attends to user emotion explicitly. With GPT~\citep{gpt} as its backbone, CAiRE~\citep{caire} adds an user-emotion-detection auxiliary objective in addition to the conventional response language modelling.~\citet{moel}, on the other hand, lined up specialised response generators that are each trained to reply to user utterances of a unique emotion.

\textbf{Desirable} or \textbf{future} emotion of the user or sentiment-aware agent has also been explored.~\citet{emp-transfo} encouraged the agent to learn an appropriate emotion for its response, whereas~\citet{emoelicitor} conditioned their chatbot utterance on the desirable user emotion that the agent is trying to elicit. Within a reinforcement learning framework,~\citet{happybot} rewarded response candidates likely to induce positive user emotion.

\section{Data}
\label{sec:data}
The datasets we have considered for our problem are introduced below. Among them, \pec/ will be used for the labelling step and \hrpmi/ for the detecting step.

\textbf{P}ersona-based \textbf{E}mpathetic \textbf{C}onversation (\pec/)~\citep{pec} features 291k one-to-one general conversations crawled from 2 subreddits: \textit{r/happy}\footnote{\url{https://www.reddit.com/r/happy/}} and \textit{r/offmychest}\footnote{\url{https://www.reddit.com/r/offmychest/}}, both considered to consist mainly of empathetic conversations, as well as another 725k one-to-one casual conversations from \textit{r/CasualConversation}\footnote{\url{https://www.reddit.com/r/CasualConversation}} used as the control group, based on the assumption that casual conversations are less empathetic. The author verified the empathy and non-empathy of these conversation corpora with the ``substantial agreement'' shown by Fleiss' kappa~\citep{fleiss-kappa} among the empathy annotators. We further group the 291K empathetic conversations as \emppec/ and the 725K non-empathetic ones as \nonemppec/.

\citet{high-low-quality-mi} created the first and, to the best of our knowledge, only publicly available dataset of MI conversations, which we term \textbf{\hlrpmi/}. The dataset comprises 259 role-play conversations collected from public video-sharing sources such as YouTube and Vimeo, with 155 labelled as high-quality counselling and the other 104 as low-quality. The conversations are available as both videos and transcripts, without any extra annotation. We denote the high-quality counselling subset as \hrpmi/ and the low-quality subset as \lrpmi/.

\section{Proposed Methodology}
\label{methdology}
Our method follows a labelling-detecting two-step setup. Both the labeller and the detector will have pre-trained language models as their underlying architectures. In particular, we will explore extending the detector with additional-input regimes and multi-task designs that can boost its performance and improve its explainability.

\subsection{Step 1: Labelling}
\label{sec:labelling}

\subsubsection{Task Definition}
\label{sec:labelling-task-definition}
The labeller, \labeller/, will be trained to determine whether a listener response is empathetic given the most recent speaker-listener talk turns. More concretely, \labeller/ will be given as input $H_{t,N} \allowbreak = \allowbreak \{u^S_{t-N+1}, \allowbreak u^L_{t-N+1}, \allowbreak \cdots, \allowbreak u^S_t, \allowbreak u^L_t\}$, a small window of the latest $N$ (e.g. $N = 2$) exchanges between the interlocutors at time step $t$ where $u^S_i$ is the $i$-th speaker utterance and $u^L_j$ is the $j$-th listener utterance, and produce a binary classification on whether $u^L_t$ is an empathetic response to $u^S_t$. Note that we provide the $2N$-utterance window to allow for more context for the labeller.

\subsubsection{Training}
\label{sec:labelling-training}
\begin{figure}[t]
  \centering
  \includegraphics[width=\linewidth]{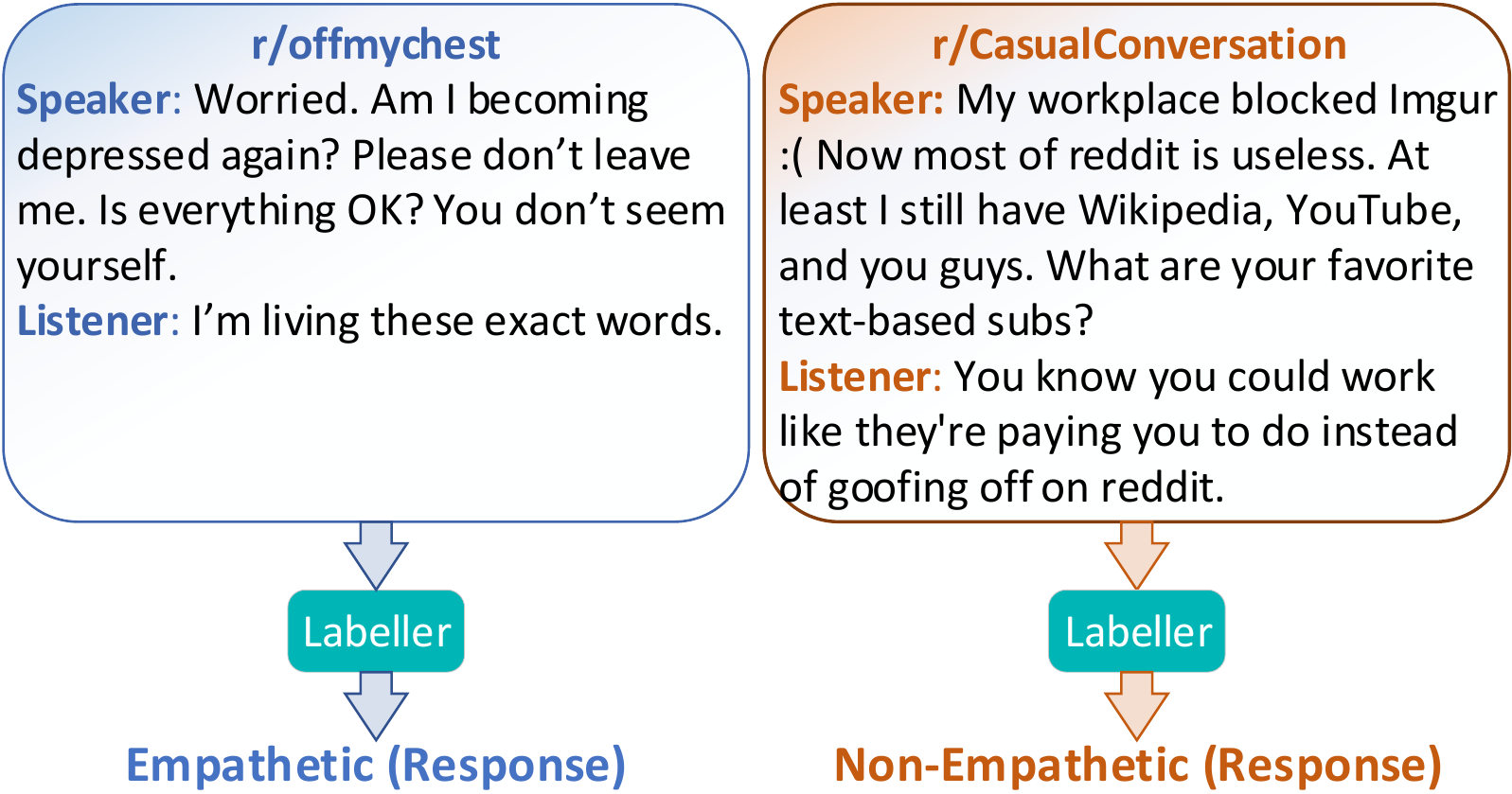}
  \caption{Training the labeller with a 2-utterance window over each \pec/ conversation. Every listener reply in subreddits \textit{r/happy} and \textit{r/offmychest} is considered an empathetic example, while every response in \textit{r/CasualConversation} is deemed non-empathetic.}
  \label{fig:training_labeller}
\end{figure}

Using \pec/ to train \labeller/, we consider the initiator of a conversation as the speaker and the other interlocutor as the listener. Every listener response and its context in \emppec/ will be deemed a positive (empathetic) example, and likewise each listener reply with its context in \nonemppec/ will be used as a negative (non-empathetic) example, as shown in Figure~\ref{fig:training_labeller}.

We will use a pre-trained language model such as BERT~\citep{bert} or GPT-2~\citep{gpt-2} as the backbone of \labeller/. As input to BERT, for example, we will represent a 4-turn conversation, assuming a window length of 4, as $\{[CLS], \allowbreak [S], \allowbreak u^S_1, \allowbreak [L], \allowbreak u^L_1, [S], \allowbreak u^S_2, \allowbreak [L], \allowbreak u^L_2\}$, where $[S]$, $[L]$, and $[CLS]$ are special tokens. Specifically, $[S]$ indicates the utterance that follows is from the speaker, the same goes for $[L]$ for the listener, while the top-layer BERT representation of $[CLS]$\footnote{$[CLS]$ is a special token required by BERT~\citep{bert} and its variants. It is placed at the beginning of the input, and its top-layer BERT representation is usually seen as a high-dimensional representation of the entire input. See~\citep{bert} for more details.} will be fed to a multi-layer perceptron (MLP) that will ultimately decide if $u^L_2$ is an empathetic response to $u^S_2$.

Once \labeller/ is trained, it will be used to annotate the MI corpus, considering the client as the speaker and the therapist as the listener, such that each therapist response in \hlrpmi/ is labelled as empathetic or non-empathetic.

\subsection{Step 2: Detection}
\label{sec:detecting}

\subsubsection{Task Definition}
\begin{figure}[t]
  \centering
  \includegraphics[width=\linewidth]{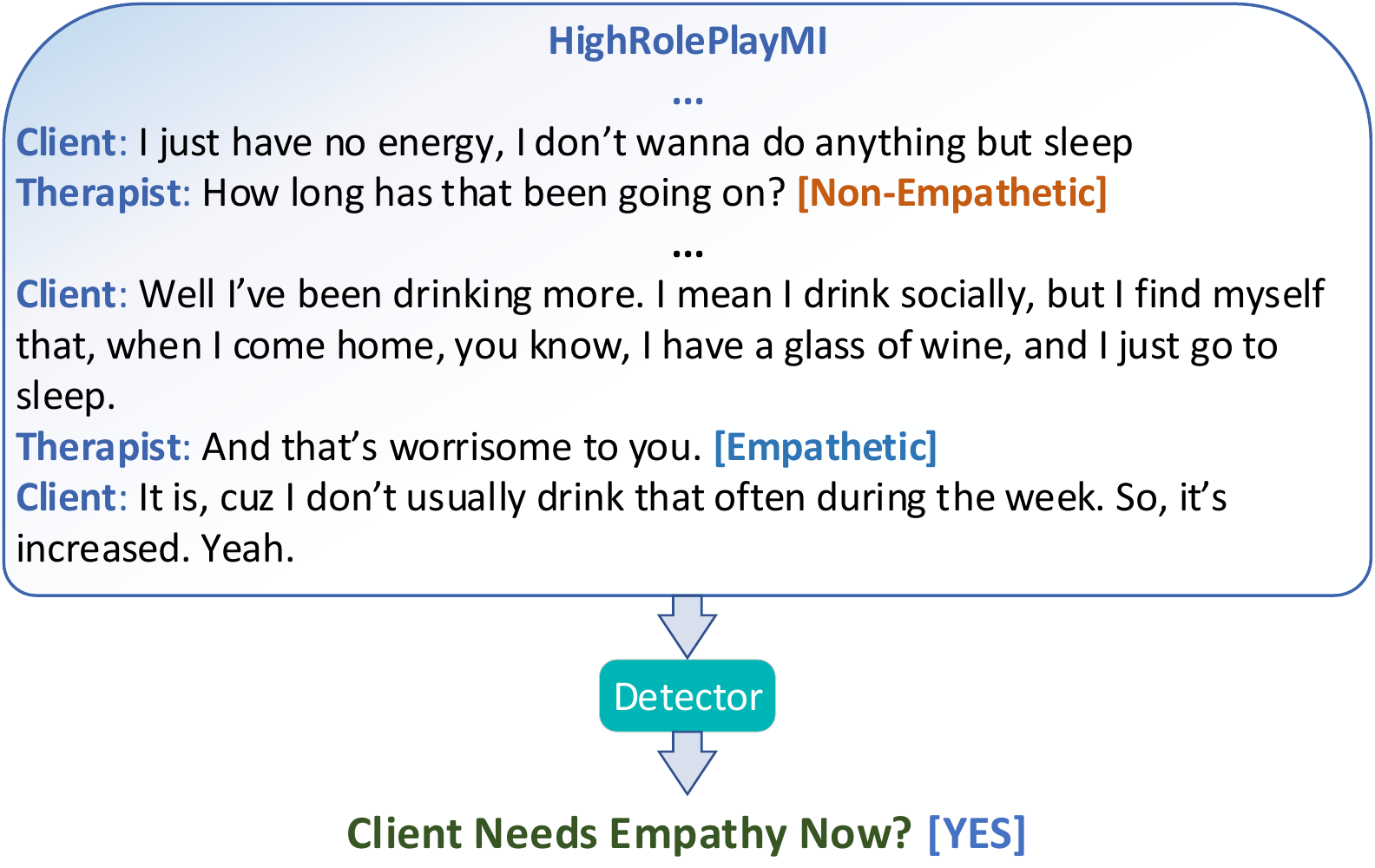}
  \caption{Training the detector with (partial) MI conversation history. The ``non-empathetic'' and ``empathetic'' labels of the therapist utterances are given by the labeller, with the client as the speaker and the therapist as the listener.}
  \label{fig:detector}
\end{figure}

Once the accuracy for Step 1 has been validated, we define the \textbf{need-for-empathy detection task} as equivalent to determining whether the therapist should show empathy in their immediate response to the client, given the history of the conversations so far. More formally, we represent the therapeutic dialogue history as $H^{thr}_t \allowbreak = \{u^C_1, \allowbreak u^T_1, \allowbreak u^C_2, \allowbreak u^T_2, \allowbreak \cdots, \allowbreak u^C_{t-1}, \allowbreak u^T_{t-1}, \allowbreak u^C_{t}\}$, where $u^C_i$ is the $i$-th client utterance and $u^T_j$ is the $j$-th therapist utterance, and the task is to decide whether $u^T_{t}$, the response to $u^C_{t}$, should be empathetic given $H^{thr}_t$, as illustrated in Figure~\ref{fig:detector}.

\subsubsection{Baseline}
\label{sec:detecting-baseline}
We will train a baseline detector \detector/ on the now-annotated \hrpmi/ to approach the task, with the input adapted from the representation described in Section~\ref{sec:labelling-training}. More formally, we will allow a much larger window of length $2M+1$ ($M \gg N$) for the most recent $2M+1$ utterances: $H^{thr}_{t,M} \allowbreak = \allowbreak \{u^C_{t-M}, \allowbreak u^T_{t-M}, \allowbreak \cdots, \allowbreak u^C_{t-1}, \allowbreak u^T_{t-1}, \allowbreak u^C_{t}\}$, and the input to the base pretrained language model, in the case of BERT~\citep{bert} or its variants (e.g. ALBERT~\citep{albert}, RoBERTa~\citep{roberta}), will be $\{[CLS], \allowbreak [C], \allowbreak u^C_{t-M}, \allowbreak [T], \allowbreak u^T_{t-M}, \allowbreak \cdots, \allowbreak [C], \allowbreak u^C_{t-1}, \allowbreak [T], \allowbreak u^T_{t-1}, \allowbreak [C], \allowbreak u^C_{t}\}$, where $\allowbreak[C]$ is the special token for the client and $[T]$ for the therapist. The MLP above the language model will then reach a binary classification using the representation of the $[CLS]$, on whether $u^T_t$ needs to be empathetic.

Considering the potential susceptibility of this system to overfitting due to the small size of \hrpmi/, we propose 2 extensions to the baseline: \textbf{emotion recognition} and \textbf{therapist response generation}, as auxiliary objectives or for additional input. Their details are explained in the remainder of this section.

\subsubsection{Extension: Emotion Recognition}
\label{sec:emotion-recognition}
Acknowledging emotions is essential to empathetic communication~\citep{facebook-empathy}. Considering that emotion recognition~\cite{DBLP:journals/cim/RecuperoABGT19} is a well-resourced task, we hypothesise that involving emotions explicitly will be beneficial for the learning process of the detector and may enable more explainability.

For \textbf{additional input} to the baseline, an external emotion classifier $cls_{emt}$ will be trained on an emotion-labelled conversation dataset, e.g. \empdial/~\citep{facebook-empathy}, and provide an emotion label prepended at the beginning of each utterance before the conversation window is fed to \detector/ as the input, similar to~\citep{facebook-empathy}.

In a \textbf{multi-task} setting, we will feed the representation of $[CLS]$ to two MLPs: one for need-for-empathy detection as in the baseline, and the other for emotion recognition. By assigning more weight to the detection loss and less to the recognition loss, we will achieve main (detection) - auxiliary (recognition) multi-task training.

\subsubsection{Extension: Therapist Response Generation}
\label{sec:therapist-response-generation}
\detector/ is effectively asked to forecast the empathy label of a therapist response absent in the input, which makes the task harder than the empathy labelling task. To facilitate the learning of the detector, we will leverage the actual therapist responses in \hrpmi/.

For \textbf{additional input} to the baseline, an external open-domain chatbot (e.g.~\citep{blender}) will be fine-tuned on some counselling (not necessarily MI) conversation datasets to adapt to the therapeutic domain, and then used to, for each input, produce response candidates that are appended to the input in the need-for-empathy detection task.

In a \textbf{multi-task} setting, we can utilise the encoder-decoder backbone of an open-domain chatbot, but the encoder output is now fed to both 1) an MLP for need-for-empathy detection, and 2) the decoder for response generation. We make the detection the main task and the response generation the auxiliary task in a joint end-to-end training regime, which may allow for insights into the correlation between the detected need for empathy and the generated response.

\section{Conclusion}
\label{sec:conclusion}
We proposed a novel task of detecting client need for empathetic therapist response. Our plan for tackling this challenge is centred around a labeller-detector design, where we 1) first train a labeller on empathy-related general conversations so that it can automatically annotate an unlabelled MI corpus with binary empathy labels, and 2) then train a detector on the newly labelled MI corpus to detect real-time client need for empathetic response. We also laid out our plans for extending the detector with additional-input and multi-task schemes to improve its ability and explainability. Expecting likely challenges in our experiments arising from the domain shift from general conversation to therapy, we hope our proposal and upcoming results will stimulate more research in empathy-related analysis for clinical counselling.

\begin{acks}
This work has been funded by the EC in the H2020 Marie Skłodowska-Curie PhilHumans project, contract no. 812882. The authors would also like to thank Dr. Mark Aloia for his guidance and support.
\end{acks}

\bibliographystyle{ACM-Reference-Format}
\bibliography{references}

\end{document}